\documentclass[manuscript]{acmart}

\AtBeginDocument{%
  }

\usepackage{listings}
\usepackage{multirow}

\begin{document}

\title{Beyond Vanilla Fine-Tuning: Leveraging Multistage, Multilingual, and Domain-Specific Methods for Low-Resource Machine Translation}


\author{Sarubi Thillainathan}
\affiliation{            
  \institution{Department of Computer Science and Engineering, University of Moratuwa}
  \city{Katubedda}
  \country{Sri Lanka} 
}

\affiliation{            
  \institution{Department of Language Science and Technology, 
Saarland Informatics Campus, Saarland University}
  \city{Saarbrücken}
  \country{Germany} 
}
\email{sarubi@coli.uni-saarland.de}
\orcid{0000-0001-8229-2497}

\author{Songchen Yuan}
\affiliation{%
  \institution{Department of Computer Science, University of Toronto}
  \city{Toronto}
  \country{Canada}}
\email{ysc.yuan@mail.utoronto.ca}

\author[label4]{En-Shiun Annie Lee}
\affiliation{%
  \institution{Department of Computer Science, University of Toronto}
  \city{Toronto}
  \country{Canada}}

\affiliation{%
  \institution{Faculty of Science, University of Ontario of Technology}
  \city{Oshawa}
  \country{Canada}}
\email{annie.lee@cs.toronto.edu}

\author[label1]{Sanath Jayasena}
\affiliation{            
  \institution{Department of Computer Science and Engineering, University of Moratuwa}
  \city{Katubedda}
  \country{Sri Lanka} 
}
\email{sanath@cse.mrt.ac.lk}

\author{Surangika Ranathunga}
\affiliation{            
  \institution{Department of Computer Science and Engineering, University of Moratuwa}
  \city{Katubedda}
  \country{Sri Lanka} 
}
\affiliation{
    \institution{School of Mathematical and Computational Sciences, Massey University}
            \city{Auckland}
            \country{New Zealand}
            } 
\email{s.ranathunga@massey.ac.nz}

\renewcommand{\shortauthors}{Thillainathan et al.}

\begin{abstract}
 Fine-tuning multilingual sequence-to-sequence large language models (msLLMs) has shown promise in developing neural machine translation (NMT) systems for low-resource languages (LRLs). However, conventional single-stage fine-tuning methods struggle in extremely low-resource NMT settings, where training data is very limited. This paper contributes to artificial intelligence by proposing two approaches for adapting msLLMs in these challenging scenarios: (1) continual pre-training (CPT), where the msLLM is further trained with domain-specific monolingual data to compensate for the under-representation of LRLs, and (2) intermediate task transfer learning (ITTL), a method that fine-tunes the msLLM with both in-domain and out-of-domain parallel data to enhance its translation capabilities across various domains and tasks. As an application in engineering, these methods are implemented in NMT systems for Sinhala, Tamil, and English (six language pairs) in domain-specific, extremely low-resource settings (datasets containing fewer than 100,000 samples). Our experiments reveal that these approaches enhance translation performance by an average of +1.47 bilingual evaluation understudy (BLEU) score compared to the standard single-stage fine-tuning baseline across all translation directions. Additionally, a multi-model ensemble further improves performance by an additional BLEU score.
\end{abstract}



\keywords{low-resource languages neural machine translation,  large language models, fine-tuning, pre-training}


\maketitle
\section{Introduction}
\label{intro}


Neural Machine Translation (NMT) systems have achieved remarkable success in recent years, with significant improvements in translating both high-resource and low-resource languages. A promising approach that has emerged to improve the translation of these diverse language pairs is the fine-tuning (single-stage) Large Language Models, particularly multilingual sequence-to-sequence Large Language Models (msLLMs) such as mBART~\citep{liu-etal-2020-multilingual-denoising,tang-etal-2021-multilingual} and mT5~\citep{xue-etal-2021-mt5}. This technique involves fine-tuning pre-trained msLLMs directly on the language pairs, leveraging their extensive pre-trained knowledge to deliver substantial performance gains compared to training an NMT system from scratch with randomly initialized weights. Studies~\citep{liu-etal-2020-multilingual-denoising,tang-etal-2021-multilingual,thillainathan2021fine,lee-etal-2022-pre} have demonstrated that this method significantly boosts translation accuracy, making it a viable option for improving NMT outcomes.




However, while single-stage fine-tuning has shown impressive results, it often falls short for Low-Resource Languages (LRLs)\footnote{To define the extent of resource scarcity, we adopt the threshold proposed by~\cite{10.1145/3567592}, classifying NMT's training data as low-resource or extremely low-resource when the available parallel corpus contains fewer than 0.5 million or 0.1 million sentences, respectively. Nonetheless, it is important to note that this threshold is not an absolute criterion.}, particularly those that are underrepresented in the pre-trained msLLMs due to limited monolingual data. This problem is more notable in domain-specific translation. As demonstrated by~\cite{lee-etal-2022-pre}, a straightforward, single-stage fine-tuning of msLLMs tends to yield suboptimal outcomes for these LRLs, requiring more advanced strategies to improve translation performance in such contexts.

Moreover, even generative LLMs, continue to struggle significantly with the translation~\cite {xu2024paradigmshiftmachinetranslation}. The performance gap between these models and msLLMs remains particularly wide in LRL settings. This underscores the ongoing need for refined training approaches that better accommodate the nuances and scarcities associated with LRL.

In this study, we aim to address the existing limitations in the performance of msLLMs, particularly in domain-specific settings,  by answering the following research questions:


\begin{enumerate}
    \item How can the translation performance of msLLMs for LRLs be enhanced by addressing the underrepresentation of these languages in the training data?
    
    
    \item Beyond conventional single-stage fine-tuning, what alternative techniques can be employed to leverage parallel corpora from different domains to enhance the adaptability and accuracy of msLLMs in translating LRLs?
    \item Considering these advanced techniques, how can we develop a comprehensive strategy to significantly improve the performance of msLLMs in translating LRLs in domain-specific settings within NMT frameworks?

\end{enumerate}


To answer the question (1) and (2), we investigate two techniques that utilize additional datasets to assist in the training process - Continual Pre-Training (CPT)\footnote{CPT—which entails further pre-training of an LLM with monolingual data, refer to Section-\ref{related_work_cpt} for more details}
and Intermediate Task Transfer Learning (ITTL)\footnote{ITTL-fine-tuning with the msLLMs with auxiliary parallel data before fine-tuning with the
limited target domain data, refer to Section-\ref{related_work_ft_msllms}}. For question (3), we combined the aforementioned CPT and ITTL two techniques to advance msLLMs for LRL-NMT in domain-specific contexts:
\begin{enumerate}
    \item \textbf{Contiually Pre-Train msLLMs} with domain-specific monolingual corpora, followed by
    \item \textbf{Fine-Tuning these CPT-enhanced msLLMs} using in- and out-domain parallel corpora with ITTL.
    We propose two ITTL fine-tuning strategies
    \begin{enumerate}
        \item a three-stage bilingual fine-tuning method, involving sequential fine-tuning with out-domain, mixed-domain, and in-domain parallel data
        \item a hybrid multilingual and bilingual fine-tuning approach, combining multilingual fine-tuning followed by bilingual fine-tuning.
    \end{enumerate}
    
    \item This methodology is further enhanced by the use of ensemble strategies to optimise results. 



\end{enumerate}

We selected Sinhala, Tamil, and English in extremely LRL domain-specific settings. Among these three languages, we experiment with the possible combination of six language pair translation directions. Our proposed CPT with ITTL strategies outperforms the baseline (direct fine-tuning) strongly for all the six directions where we obtained  $+2.6$, $+1.83$, $+1.41$, $+1.17$, $+1.03$ and $+0.8$ BLEU score improvement respectively for  $Ta\rightarrow En$,   $Ta\rightarrow Si$,  $Si\rightarrow Ta$,  $Si\rightarrow En$,  $En\rightarrow Ta$ over strong direct fine-tuning baseline. Furthermore, our multi-model ensemble improves the performance even further by an average of (+2.13) BLEU score.

\section{Methodology} 
\subsection{Overview}
\label{metho:overview}


\begin{figure*}[htp]
    \centering
    \includegraphics[scale=0.291]{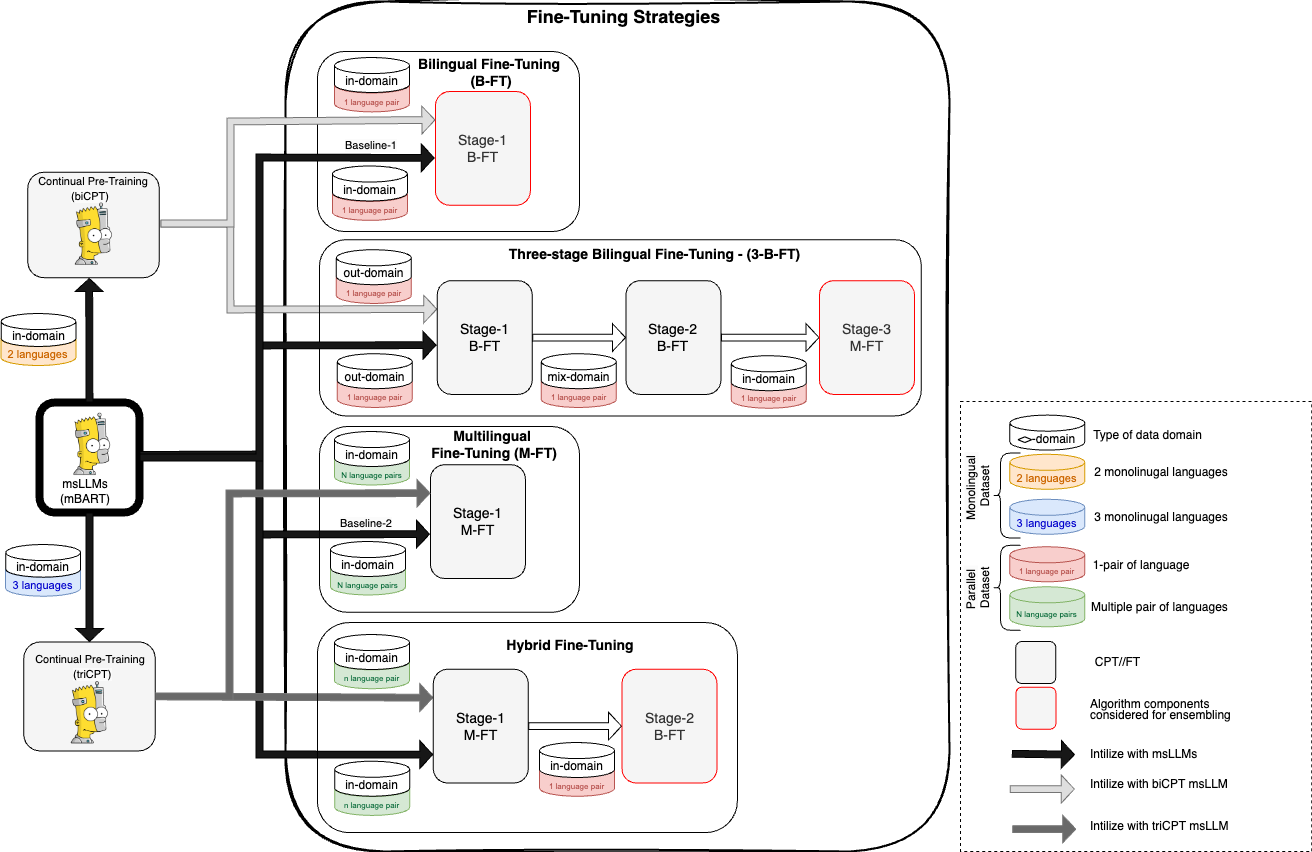}
    \caption{Overview of Methodology}
    \label{fig:methodology}
\end{figure*}

We use two single-stage fine-tuning baselines- Bilingual Fine-Tuning (B-FT) and Multilingual Fine-Tuning (M-FT) as described in~\cite{tang-etal-2021-multilingual}. In B-FT we fine-tune one language pair at a time with domain-specific parallel data. Meanwhile, in M-FT, we fine-tune more than one language pair simultaneously with domain-specific parallel data. M-FT is further categorized into three types: 
\begin{enumerate}
    \item Many-to-One Multilingual Fine-Tuning (M2O-FT): $Many \rightarrow Language$ \textit{X}
    \item One-to-Many Multilingual Fine-Tuning (O2M-FT): \textit{Language} $X \rightarrow Many$
    \item Many-to-Many Multilingual Fine-Tuning (M2M-FT): $Many\rightarrow Many$, with Language X as a pivot language.
\end{enumerate}

In both B-FT and M-FT, training of the NMT model starts with the weights of the msLLM instead of random initialization. In other words, both the encoder and the decoder of the NMT model are initialized with the weights of the msLLM. The overview of this fine-tuning process is described in Fig.~\ref{fig:methodology}.

We investigate CPT and ITTL to obtain better results over single-stage fine-tuning of msLLMs for domain-specific LRL-NMT. These are the main three approaches:

\begin{itemize}
    \item Continual Pre-Training (CPT) of the msLLM for domain adaptation.

    \item ITTL fine-tuning of msLLM with in- and out-domain parallel data for the NMT task.

    \item Ensembling different fine-tuned models.

\end{itemize}
 Fig.~\ref{fig:methodology} summarises how our techniques are implemented/related. 

\subsection{Continual Pre-training for Domain Adaptation}
\label{sec:metho_cpt}



To mitigate the under-representation of LRLs within the msLLM, we extend the pre-training phase of the msLLM by integrating supplementary monolingual data. During this CPT process, the msLLM is exposed to a procedure wherein the monolingual data is deliberately `noised', or altered, and then presented as the input (source). The model is then tasked with reconstructing or translating this noised input back into its original, unaltered form (target), a method known as self-supervised learning.

Our approach to CPT encompasses the utilization of various types of monolingual data, detailed as follows:

\begin{itemize}
\setlength{\itemindent}{4em}
    \item [Case A:] Only with in-domain monolingual data.
    
    \item [Case B:] Only with out-domain monolingual data, which is larger in quantity. 
    
    \item [Case C:] By combining both in-domain and out-domain monolingual data in two ways. 
    \begin{itemize}
        \setlength{\itemindent}{8em}
        \item [Case C1:]  Mix in-domain CPT: Out-domain data together with in-domain data and create a mixed-domain dataset and then pre-train. 
        \item [Case C2:] Sequential CPT: Pre-training first with large out-domain and then in-domain data. 
    \end{itemize}
    
\end{itemize}

\subsection{Intermediate Task Transfer Learning (ITTL)}
\label{sec:metho_ITTL_ft}

We proposed two different ITTL fine-tuning strategies as follows: (exploration are visually represented and elaborated upon in Fig.~\ref{fig:methodology})   

\begin{enumerate}
    \item Three-stage Bilingual Fine-tuning [3-B-FT]: 
    We borrowed the idea of mix-fine-tuning from~\cite{chu-etal-2017-empirical}, where they first train a Transformer-based NMT model from scratch with out-domain parallel corpus and then fine-tune it on a parallel corpus that is a mix of the in-domain and out-domain corpora. However, they did not use any msLLMs.

    In our study, we propose the following on msLLMs:
    
    \begin{enumerate}
        \item Initialize training with msLLM.
        \item B-FT with out-domain parallel data. 
        
        \item B-FT with mixed domain parallel data where we up-sampled the in-domain data size to match with the larger out-domain corpus and created the mixed domain parallel data.
        \item Finally, B-FT with in-domain parallel data.
    \end{enumerate}

    \item Multilingual Fine-tuning followed with Bilingual Fine-tuning [M-FT,B-FT]:
     \begin{enumerate}
        \item Initialize training with the msLLM.
        \item Conduct M-FT using in-domain parallel data. This could be O2M-FT, M2O-FT, or M2M-FT, and select the best-performing model.
        \item Finally, tailer to one particular translation direction by doing  B-FT with in-domain parallel data.
    \end{enumerate}
    
\end{enumerate} 

\subsection{Ensemble of Fine-tuned msLLMs}
\label{sec:metho_ensemble}

As we propose multiple CPT and ITTL techniques to improve msLLMs for domain-specific NMT, we utilize multiple similar models that share identical target vocabularies and the same decoding parameters; all are based on the same foundational msLLMs. To further enhance performance, we explore the potential of ensembling multiple fine-tuned msLLMs.

Ensembling is a machine learning technique that combines several base models to create an optimized predictive model~\citep{hansen1990neural}. In the context of NMT, ensembling involves translating the input sentence using multiple models, and then averaging the outputs from each to produce the final translation. We implement primarily two techniques for ensembling:

\begin{enumerate}
    \item Checkpoint Ensemble
    
    This is the bare minimum way to conduct the ensemble from a single training process~\citep{Chen2017CheckpointEE}. Previous studies applied checkpoint ensembling on RNN based NMT systems by combining the last N checkpoints of a single training~\citep{sennrich2016edinburgh,sennrich-etal-2017-university,imamura-sumita-2017-ensemble}. In this study, we apply this checkpoint ensembling on msLLMs. We combine 3 different saved checkpoints from a particular single training. In other words, when we are fine-tuning the msLLM, we save the last ten checkpoints. We select 3 checkpoints that yield the best results in the validation test set among all the saved checkpoints. Then, we apply the checkpoint ensembling using these selected 3 checkpoints. In addition, we also experiment with the combination of 2 checkpoints. Finally, we report the best ensemble result among both of the above scenarios. Note that we can combine any number of models we want.
    
    \item Multi-model Ensemble
    
    Besides our baseline (single-stage fine-tuning), we explore different techniques to improve the fine-tuning process of the msLLM (for example [3-B-FT] as shown in Fig.~\ref{fig:methodology}), resulting in multiple similar msLLMs that have been fine-tuned with in-domain parallel data at the final stage. We combine a maximum of 3 such msLLM fine-tuned models (due to our computational resource limitations) for the ensemble. We pick the top 3 best models out of these different approaches. 
\end{enumerate}

\section{Implementation}

\subsection{Dataset}

For our experiments, we pick two languages that are underrepresented in the mBART model: Sinhala (Si) and Tamil (Ta), along with English (En). 

\textbf{Domain-specific Parallel data} \\ 
Our domain-specific parallel datasets are from official government documents of Sri Lanka~\citep{Fernando2020DataAA}, consisting of annual reports, crawled contents from government institutional websites, committee reports, procurement documents and acts. According to the statistics of our dataset given in Table~\ref{table:parallel-dataset-details}, our dataset is smaller than 100k. Thus this creates an extremely low-resource domain-specific translation task.

\textbf{Out-domain Parallel data} \\
We also gather publicly available out-domain parallel datasets from OPUS\footnote{\url{https://opus.nlpl.eu/}} and WMT MT tasks\footnote{\url{https://www.statmt.org/wmt20/translation-task.html}}. For Sinhala-English, we use the FLoRes V1 training dataset\footnote{\url{https://github.com/facebookresearch/flores/tree/main/floresv1}}~\citep{guzman-etal-2019-flores}. For Tamil-English, we use WMT20\footnote{\url{https://www.statmt.org/wmt20/translation-task.html}} MT news tasks' parallel datasets. For Sinhala-Tamil we use NLLB dataset from~\cite{costa2022no}. From NLLB, we pick the top 100,0000 samples after sentence similarity storing. Statistics of the dataset are given in Table~\ref{table:parallel-dataset-details}.

\textbf{Monolingual data for CPT} \\
For monolingual data, we use domain-specific monolingual data obtained from~\cite{epaliyana2021improving} for our selected languages. For out-domain monolingual data, we use the online news site NewsFirst\footnote{\url{https://www.newsfirst.lk/}}~\citep{rajitha2020sinhala} and FLoRes\footnote{\url{https://github.com/facebookresearch/flores/tree/main/floresv1}}~\citep{guzman-etal-2019-flores}. The statics are given in Table~\ref{table:mono_data}.

\begin{table*}[htbp]
\begin{center}
\begin{tabular}{| l | l| l | c | c |}
\hline
\textbf{Domain} & \textbf{Dataset} & \textbf{Language pair} & \textbf{No. of Sentence}\\
\hline
\multirow{5}{*}{In-Domain} & \multirow{5}{*}{Government}  & Sinhala-Tamil (train) &  66,348 \\
\cline{3-4}
 & & Tamil-English (train) &  66,348\\
\cline{3-4}
 & & Sinhala-English (train)   & 74,468\\
\cline{3-4}
 & & for all pairs (validation)  & 1,623\\
\cline{3-4}
& & for all pairs (test) & 1,603\\
\hline
\multirow{3}{*}{Out-Domain} & FLoRes & Sinhala-English  (train)  & 646,781\\
\cline{2-4}
 &  WMT20 News & Tamil-English  (train) &  305,671\\
\cline{2-4}
& NLLB & Sinhala-Tamil  (train)  &  100,000\\
\hline
\end{tabular}
\caption{Statistics of the parallel dataset. In-domain dataset obtained from official government documents~\citep{Fernando2020DataAA,fonseka2020english}}
\label{table:parallel-dataset-details}
\end{center}
\end{table*}

\begin{table*}[htbp]
\begin{center}
\begin{tabular}{|c|c|c|c|}
\hline
\textbf{Domain} &\textbf{Dataset} &  \textbf{Language}  &\textbf{No. of Sentences}  \\
\hline
 \multirow{3}{*}{In-domain} &  \multirow{3}{*}{Government Data} & Sinhala & 44,115 \\
\cline{3-4}
& &  English  & 42,773 \\
\cline{3-4}
& &  Tamil &  24,220 \\
\hline
\multirow{4}{*}{Out-domain}  &   \multirow{2}{*}{News-first} &  Sinhala & 650,000 \\
\cline{3-4}
 && English  & 627,301 \\
\cline{2-4}
& \multirow{2}{*}{FLoRes} &  Sinhala  & 646,781 \\
\cline{3-4}
 & & English  & 646,781 \\
\hline
\end{tabular}
\caption{Statistics of the monolingual dataset.}
\label{table:mono_data}
\end{center}
\end{table*}

\subsection{msLLM selection}

For our experiments, we select mBART50. It supports all the considered languages. In particular, mBART has shown promising results for supervised and unsupervised NMT~\citep{liu-etal-2020-multilingual-denoising,tang-etal-2021-multilingual}. mBART is memory efficient and has shown relatively better results than mT5~\citep{lee-etal-2022-pre}. We follow the same mBART model architecture --- the standard sequence-to-sequence Transformer~\citep{10.5555/3295222.3295349}, with 12 layers of encoder-decoder with the model dimension of 1024 on 16 heads. For training, we use the FairSeq\footnote{\url{https://github.com/pytorch/fairseq}} tool.





\subsection{Baselines} 
\label{sec:imple_baseline}
As mentioned in Section~\ref{metho:overview}, we consider Bilingual Fine-tuning (B-FT) and Multilingual Fine-Tuning (M-FT) as our baselines. For all directions, we fine-tune parallel data with 0.3 dropout, 0.2 label smoothing, 2500 warm-up steps, 3e-5 maximum learning rate as described by~\cite{liu-etal-2020-multilingual-denoising}. We use a maximum of up to 100k training updates for B-FT case. For the M-FT case, we applied the same setting as B-FT and used a maximum of 300k training updates. Additionally, we applied a temperature sampling rate of 1.5. Since our main goal is to conduct M-FT in a non-English centric manner, we chose Sinhala centric M-FT as well as Tamil centric M-FT. It is the most requested use case in Sri Lanka since the government documents are mostly produced in Sinhala and should be translated to Tamil and English~\citep{farhath2018integration}. Non-English translation, of course, has been identified as a common requirement in general for MT~\citep{10.5555/3546258.3546365}.

\subsection{Continual Pre-training for Domain Adaptation}

We explore different data selections for the CPT strategies proposed in Section~\ref{sec:metho_cpt} as described below:

\begin{itemize}
    \item [Case A:] CPT with in-domain monolingual data.
    
    We use available in-domain government monolingual data (see Table~\ref{table:mono_data}). However, the monolingual dataset is small - 44k, 42k, 24k for Sinhala, English and Tamil respectively. 
    
    Next, we utilize parallel data (as shown in Table~\ref{table:parallel-dataset-details}) as monolingual data for each selected language pair and the available small monolingual government data to create a combined dataset. First, we use only the selected two languages for pre-training (biCPT). Then we move to CPT with three languages for M-FT, which we refer to as triCPT

    \item [Case B:] CPT with out-domain monolingual data, which is larger in quantity. 
    
    We use the much larger out-domain monolingual data mainly obtained from news data crawled from Sri Lankan news websites~\citep{rajitha2020sinhala} and      FLoRes (see Table \ref{table:mono_data}). 
    
    \item [Case C:] CPT by combining both in-domain and out-domain monolingual data. 
    \begin{itemize}
        \item [Case C1:] CPT with mixed domain (in-domain + out-domain) monolingual data. 
        For this case, we took the FLoRes~\citep{guzman-etal-2019-flores} out-domain monolingual data and mixed it with our in-domain government monolingual dataset mentioned in Case A. 
        
        \item [Case C2:] Multistage CPT on large out-domain and then in-domain data. 
        Here we first pre-train with large out-domain monolingual data (case B) and then pre-train with in-domain monolingual data (case A)
    \end{itemize}
    
\end{itemize}

\textbf{Pre-training (denoising) mBART50 msLLM} \\
We further pre-train the mBART50 msLLM using the same objective function used for the initial pre-training of the model. 

 Our CPT data covers $K$ languages: $D = {D_1, ..., D_K}$, 
where each $D_i$ is monolingual data in a language $i$. 
Equation~\ref{eq:learning} is the noising function$g$ that corrupts the text~\citep{liu-etal-2020-multilingual-denoising}. We train the model to predict the original text $X$ given $g(X)$. More formally, we aim to maximize $L_\theta$:

\begin{equation}
L_{\theta}= \sum_{D_{i}\in{D}} \sum_{X\in{D_{i}}} log P(x|{g}(x);\theta)
\label{eq:learning}
\end{equation}

\begin{itemize}
    \item where $X$ is an instance in language $i$
    \item the distribution $P$ is defined by the msLLM.
\end{itemize}

We use the noising techniques used by the mBART model: Random Span masking and Order Permutation~\citep{liu-etal-2020-multilingual-denoising}. First, we select spans of text and replace it with a mask token. We mask 0.3\% of words in each instance (with random masking 0.1) by randomly sampling a span length according to a Poisson distribution ($\lambda$ = 3.5). Also, we shuffle the order of sentences within each instance. The decoder input is the original text with one position offset. A language id symbol $<LID>$ is used as the initial token to predict the sentence as in~\citep{liu-etal-2020-multilingual-denoising}.

\subsection{ITTL Fine-tuning}

We use two base models (mBART50 and best-performed CPT mBART50 which is an in-domain CPT model) to initialize the ITTL fine-tuning. We have two different ITTL fine-tuning techniques: three-stage fine-tuning and M-FT followed by B-FT as described in Section~\ref{sec:metho_ITTL_ft}. 



\subsection{Preprocessing}

\label{sec:preprocessing}
The government corpus has been cleaned and verified manually with the help of professional translators~\citep{fonseka2020english}. Sentences containing only dates, special characters and numbers have been removed. Cleaning scripts of the Moses\footnote{\url{http://www.statmt.org/moses/}}~\citep{koehn-etal-2007-moses} tool  were used to remove misaligned sentences. English sentences were tokenized using Moses toolkit\footnote{\url{http://www.statmt.org/moses/}}~\citep{koehn-etal-2007-moses}, while a tokenizer developed by~\cite{farhath2018integration} was used for Sinhala and Tamil. We use the SentencePiece\footnote{\url{https://github.com/google/sentencepiece}} model learned over monolingual Common Crawl (CC) data in the mBART~\citep{tang-etal-2021-multilingual} model, containing 250,000 sub-word tokens.

\subsection{Addressing the Zero Width Joiner (ZWJ) issue}
As mentioned earlier in Section~\ref{sec:preprocessing}, we use the SentencePiece\footnote{\url{https://github.com/google/sentencepiece}} tokenizer. However, empty tokens where Unicode appears as space in output vocabulary have been removed while training the SentencePiece model. Due to that, the Zero Width Joiner character (U+200D) has been replaced by whitespace. Hence language scripts that require Zero Width Joiner get altered and result in a wrong output. This issue is there for languages like Sinhala, Kannada and Malayalam\footnote{\url{https://en.wikipedia.org/wiki/Zero-width_joiner}}. Ideally, we should not replace the Zero Width Joiner (U+200D) with whitespace since it indicates to join two chars without zero width (no whitespace). Also, the zero-width joiner must be present to decode the segmentation of decoder outputs to raw text successfully. These special characters should be kept as they are while learning the SentencePiece model. We resolved this issue and the fix was successfully merged with SentencePiece\footnote{\url{https://github.com/google/sentencepiece/pull/630}} to eliminate this ZWJ issue.

However, we cannot eliminate this issue when using the already trained (pre-trained) models as the models were trained without ZWJ over the large-scale monolingual data. Hence we propose a post-processing solution to add the ZWJ character in the possible occurring places for Sinhala languages\footnote{\url{https://en.wikipedia.org/wiki/Sinhala_script\#Consonant_conjuncts}}. Here we mainly cover frequently occurring \textit{``yansaya"} and \textit{``rakāransaya"} of the Sinhala script. Further details of the post-processing logic are provided in~\ref{postprocessing_logic}.

\subsection{Evaluation setup}
The final model is selected based on the validation likelihood. We use beam size 5 for decoding as used by~\cite{liu-etal-2020-multilingual-denoising}. The final results are calculated against the true-target tokenized data and reported in BLEU~\citep{papineni-etal-2002-bleu}\footnote{script from ~\citep{koehn-etal-2007-moses}}

For ensembling cases described in Section~\ref{sec:metho_ensemble}, we combine a maximum of 3 models and evaluate under the same decoding and BLEU scores calculation as mentioned above.

\section{Results}
\subsection{Comparison Between Full Precision Fine-Tuning and Mixed-Precision Fine-Tuning} 

When training a neural network, usually we use Full Precision, a 32-bit floating-point (FP32) arithmetic calculation by default. Alternatively, we can use Mixed-Precision training, which combines single-precision (FP32) with half-precision (FP16) format. Mixed-Precision training has further performance benefits on NVIDIA GPUs. It requires shorter training time and lower memory requirements, enabling larger batch sizes, models, and inputs\footnote{\url{https://pytorch.org/blog/accelerating-training-on-nvidia-gpus-with-pytorch-automatic-mixed-precision/}}.

Hence, for bilingual fine-tuning (baseline), we fine-tuned each pair of directions two times under the same hyper-parameters, one with the default setting Full Precision and the second one with Mixed-precision training. As shown in Table~\ref{table:fp16}, there is not much BLEU score difference between Full Precision and Mixed-precision. However, we noticed that Mixed-precision training reduces the training time. Considering our resource limitation for training, we choose Mixed-precision training going forward.

\begin{table}[htbp]
\centering
\small
\setlength\tabcolsep{2.5pt}
\begin{tabular}{|l|c|c|c|c|c|c|}
\hline
 Models  &$Si \rightarrow Ta$ &$Ta \rightarrow Si$ & $Si \rightarrow En$ & $En \rightarrow Si$ & $Ta \rightarrow En$ & $En \rightarrow Ta$\\
\hline
Bilingual full precision FT	& 29.82 &	37.87 &	37.84 &	35.89 &	33.63 &	25.96\\
\hline
Bilingual mixed precision FT	& 29.75 &	37.57 &	37.72 &	36.11 &	33.36 &	25.8\\
\hline
\end{tabular}
\caption{Comparison between full precision training and mixed precision Fine-Tuning.}
\label{table:fp16}
\end{table} 

\subsection{Continual Pre-training (CPT) for Domain Adaptation}

Extensive evaluation results of the CPT  specifically focusing on the $Si\leftrightarrow En$ language pair are detailed in Table~\ref{table:cpt}. Overall, CPT improves the results compared to both baselines. This clearly illustrates that under-represented languages require additional pre-training to learn a good representation in the msLLM. Among the several CPT techniques we proposed, we observe the highest improvement on Case A(ii), where we get +0.79 and +0.53 on $Si \rightarrow En$ and $En \rightarrow Si$, respectively. Our experiments show that in-domain pre-training (Case A) plays a pivotal role compared to other pre-training strategies that utilize out-domain data.  Even though we do not have adequate in-domain data, our experiments validate that having little in-domain data would have more impact than out-domain or mixed-domain data. Hence, we pick CPT on in-domain monolingual data in going forward. 

When we look at biCPT+B-FT (raw 15) in Table~\ref{table:final_all_6}, we can evidence that CPT helps the NMT model to perform better on extremely LRL pairs for all six directions, where we obtained a maximum of +0.97 BLEU improvement on $Si \rightarrow Ta$.

\begin{table}[htbp]
\centering
\small
\setlength\tabcolsep{2.5pt}
\begin{tabular}{|c|l|c|c|}
\hline
 &\textbf{Models} &\textbf{$Si\rightarrow En$} &\textbf{$En\rightarrow Si$}\\
\hline
&baseline (B-FT)    &	37.72 &	36.11 \\
\hline
 \multirow{2}{*}{Case A} & (i) biCPT in-domain mono data then B-FT	&	38.21 (+0.49) &	36.38 (+0.27)\\
&(ii) biCPT combined in-domain parallel and mono data then B-FT	&	\textbf{38.51} (+0.79) &	\textbf{36.64} (+0.53)\\
\hline
Case B &biCPT out-domain mono data then B-FT	 &	38.09 (+0.37) & 36.21 (+0.1)\\
\hline
 \multirow{2}{*}{Case C} &(i) mixed in-domain and out-domain mono biCPT then B-FT   &	38.3 (+0.58) &	36.33(+0.22)\\
\cline{2-4}
 &(ii) multistage biCPT then B-FT &	38.38 (+0.66) &	36 (-0.11)\\
\hline
\end{tabular}
\caption{Bilingual fine-tuning on top of CPT models against bilingual fine-tuning on original msLLM model (mBART) for $Si\leftrightarrow En$ pairs}
\label{table:cpt}
\end{table} 

\subsection{Multilingual Fine-tuning using the msLLM}
\label{sec:Multilingual_Fine_Tuning}

The performance results of the M-FT  baseline in comparison to the B-FT baseline are presented in the first ten rows of Table~\ref{table:final_all_6}. Predominantly, the results indicate a marginal decrement in the performance of the M2M-FT relative to the B-FT baseline. This trend is postulated to stem from the insufficiency of parallel data required for training robust Multilingual NMT (MNMT) models.

Despite the general decline, overall an incremental enhancement in performance is observed for O2M-FT and M2O-FT configurations compared to M2M-FT (expect 3 scenarios). This increment aligns with previous English-centric MNMT results reported by others~\citep{johnson-etal-2017-googles,tang-etal-2021-multilingual,Arivazhagan2019MassivelyMN,aharoni-etal-2019-massively}. In our experiments, $Si-Ta$ acts as an LRL pair compared to the $Si-En$ pair, as we have around 66k and 74k sentences, respectively. When we up-sample $Si-Ta$ to match the size of $Si-En$, we were able to maximize the performance of O2M-FT and M2O-FT and beat the B-FT baseline of $Si\rightarrow Ta$, $Ta \rightarrow Si$ by +0.55 and +0.65, respectively.  
Similarly, we observe M2M-FT cases overall perform lower than [triCPT,O2M-FT], [triCPT,M2O-FT] cases. We obtained a maximum of +1.22 BLEU score improvement on triCPT M-FT models for $Ta \rightarrow Si$. 
\subsection{ITTL Fine-tuning}

Results of our two ITTL strategies [biCPT,3-B-FT] and [triCPT, M-FT,B-FT] \footnote{biCPT where we pre-train using monolingual data of two languages and whereas triCPT with three languages. biCPT will follow with B-FT and triCPT follow with M-FT as we try to fine-tune using three languages} are given in the last four raws of Table~\ref{table:final_all_6}. Three-stage fine-tuning ITTL [biCPT,3-B-FT] significantly improves BLEU score compared to baseline B-FT for all the directions where we obtained a maximum of +1.81 score improvement for $Si \rightarrow Ta$ and a minimum of +0.65 improvements on $En \rightarrow Ta$. 

Our biCPT 3-stage fine-tuning results prove that utilizing the available in-domain monolingual and out-domain parallel data enhances fine-tuning performance in extremely low-resource domain-specific settings rather than only utilizing out-domain parallel data. 

\begin{table}[htbp]
\centering
\resizebox{\columnwidth}{!} {%
\small
\setlength\tabcolsep{2.5pt}
\begin{tabular}{|l|c|c|c|c|c|c|c|}
\hline
 Models & Pivot Lang  &$Si \rightarrow Ta$ &$Ta \rightarrow Si$ & $Si \rightarrow En$ & $En \rightarrow Si$ & $Ta \rightarrow En$ & $En \rightarrow Ta$\\
   & in M-FT  & & &  &  &  & \\
\hline
B-FT (Baseline)  &  -      & 29.75         & 37.57         & 37.72         & 36.11         & 33.36          & 25.8\\
\hline
O2M-FT (Baseline) & \multirow{3}{*}{Si}	& 30.3 (+0.55)  & N/A        & 37.59 (-0.13) & N/A & N/A & N/A \\
M2O-FT (Baseline) &	& N/A          & 37.62 (+0.05) & N/A          & 36.03 (-0.08) & N/A & N/A \\
M2M-FT (Baseline) &	& 28.44 (-1.31) & 36.38 (-1.19) & 36.05 (-1.67) & 34.93 (-1.18) & N/A & N/A \\
\hline
O2M-FT (Baseline) & \multirow{3}{*}{Ta}	& N/A  & 38.22 (+0.65)    & N/A & N/A & 33.45 (+0.09) & N/A \\
M2O-FT (Baseline) &	& 29.78 (+0.03)   & N/A   & N/A & N/A & N/A & 26.27 (+0.47) \\
M2M-FT (Baseline) &	& 28.85  (-0.9)  & 36.35 (-1.22)     & N/A & N/A & 32.64 (-0.72) & 24.65 (-1.15) \\
\hline 
O2M-FT (Baseline) & \multirow{3}{*}{En}	& N/A  & N/A  & N/A  & 35.58 (-0.53)   & N/A   & 26.24 (+0.44) \\
M2O-FT (Baseline) &	& N/A  & N/A  & 37.75  (+0.03) & N/A   & 34.85  (+1.49)    & N/A \\
M2M-FT (Baseline) &	& N/A & N/A   & 36.61 (-1.11)  & 34.48  (-1.63)   & 33.99  (0.63)   & 24.68 (-1.12) \\

\hline \hline
M-FT(best),B-FT & \multirow{1}{*}{Si} & 30.97 (+1.22) &	37.54 (-0.03) &	36.1 (-1.62)	& 35.88 (-0.23) & N/A & N/A \\
\hline
M-FT(best),B-FT & \multirow{1}{*}{Ta} & 29.85 (+0.1) &	39.17	(+1.6) & N/A & N/A & \textbf{35.96	(+2.6)} & 26.22 (+0.42) \\
\hline
M-FT(best),B-FT & \multirow{1}{*}{En} & 	 N/A & N/A & 36.87 (-0.85) & 36.42 (+0.31) & 33.94 (+0.58) & 26.12 (+0.32) \\
\hline \hline
3-B-FT  & - & 30.44 (+0.69)	& 36.97 (-0.6) &	38.27 (+0.55) &	35.68 (-0.43) &	33.54 (+0.18) & 25.83  (+0.03) \\ 

\hline \hline \hline
biCPT,B-FT     & -  & 30.72 (+0.97) & 38.08 (+0.51) & 38.51 (+0.79) & 36.64 (+0.53) & 34.29 (+0.93) & 26.3 (+0.5)\\
\hline
triCPT,O2M-FT & \multirow{3}{*}{Si}  & 30.5 (+0.75) & N/A & 37.93 (+0.21) & N/A & N/A & N/A \\
triCPT,M2O-FT &  & N/A  & 38.04 (+0.47) & N/A & 35.87 (-0.24) & N/A & N/A  \\
triCPT,M2M-FT	& & 29.64 (-0.11)	& 37.15 (-0.42) &	36.98 (-0.74)	& 34.51 (-1.6) & N/A & N/A \\
\hline
triCPT,O2M-FT  &  \multirow{3}{*}{Ta} 	& N/A  & 38.79  (+1.22)  & N/A & N/A & 33.95 (+0.59) & N/A \\
triCPT,M2O-FT  & 	& 29.31  (-0.44)  & N/A   & N/A & N/A & N/A & 26.43 (+0.63) \\
triCPT,M2M-FT  &	& 28.87 (-0.88)    & 36.46  (-1.11)   & N/A & N/A & 32.61 (-0.75) & 24.25 (-1.55) \\
\hline 
triCPT,O2M-FT &  \multirow{3}{*}{En}	& N/A  & N/A  & N/A  & 35.17 (-0.94)   & N/A   & 26.63 (+0.83) \\
triCPT,M2O-FT  &	& N/A  & N/A  & 37.98 (+0.26)  & N/A   & 35.23  (+1.87)    & N/A \\
triCPT,M2M-FT  &	& N/A & N/A   & 36.85 (-0.87)  &35.74    (-0.37)  & 33.1  (-0.26)    & 25.37 (-0.43) \\
\hline 
\hline
triCPT,M-FT(best),B-FT  & Si & \textbf{31.16 (+1.41)} &	38.9 (+1.33) &	36.53 (-1.19) &	36.17 (+0.06) & N/A & N/A \\
\hline

triCPT,M-FT(best),B-FT & Ta & 30.19 (+0.44)	& \textbf{39.4 (+1.83)} & N/A & N/A & 35.15 (+1.79) & \textbf{26.83 (+1.03)} \\
\hline

triCPT,M-FT(best),B-FT & En &  N/A & N/A&  37.2 (-0.52) & 36.68 (+0.57) & 34.52 (+1.16)	 & 26.75 (+0.95) \\

\hline \hline

\hline
biCPT,3-B-FT  & -  & 30.87 (+1.12) &	39.38 (+1.81) & \textbf{38.89 (+1.17)}  & \textbf{36.91 (+0.8)} & 34.82 (+1.46) & 26.45 (+0.65)  \\

\hline 

\hline
\end{tabular}
}
\caption{CPT and ITTL against our strong single-stage fine-tuning baseline (B-FT, M-FT) - In M-FT, not applicable cases denotes with N/A. For example Si centric O2M-FT includes $Si\rightarrow Ta$ and $Si\rightarrow En$ directions and rest of the cases indicated with N/A.}
\label{table:final_all_6}
\end{table} 

Our second ITTL approach, Fine-tuning from Multilingual Fine-tuned MNMT models [triCPT,M-FT,B-FT] shows quantifiable improvements except on $Si \rightarrow En$. For $Si \rightarrow Ta$ and $Ta \rightarrow Si$, we obtain +1.41 and +1.83 BLEU score gains respectively, and also notably +1.03 on $En \rightarrow Ta$ direction.

Overall, we prove that utilizing additional in-domain monolingual data for pre-training and in- and out-domain data fine-tuning improves the results on under-represented languages in the msLLM. As observed in our initial B-FT experiments (baseline) results, here also we can witness that, even though we fine-tune with large out-domain parallel data, we still require a reasonable amount of in-domain parallel corpus to learn a good language representation on a particular domain for morphologically rich languages.

\subsection{Ensemble}

\begin{table}[htbp]
\centering
\resizebox{\columnwidth}{!} {%
\small
\setlength\tabcolsep{2.5pt}
\begin{tabular}{|c|c|c|c|c|c|c| }
\hline
 Rank  &$Si \rightarrow Ta$ &$Ta \rightarrow Si$ & $Si \rightarrow En$ & $En \rightarrow Si$ & $Ta \rightarrow En$ & $En \rightarrow Ta$\\
\hline
1 & triCPT+M-FT+B-FT (Si) & triCPT+M-FT+B-FT (Ta) & biCPT+3-B-FT & biCPT+3-B-FT & mbart50+M-FT+B-FT (Ta) & triCPT+M-FT+B-FT (Ta) \\
\hline
2 & mbart50 + M-FT + B-FT (Si) & biCPT+3-B-FT & biCPT+B-FT  & triCPT+M-FT+B-FT (En) & triCPT+M2O-M-FT (En) & triCPT+M-FT+B-FT (En) \\
\hline
3 & biCPT+3-B-FT & mbart50+M-FT+B-FT (Ta) & mBART50+3-B-FT & biCPT+B-FT & triCPT+M-FT+B-FT (Ta) & triCPT+O2M-M-FT (En)\\

\hline
\end{tabular}
}
\caption{Top 3 improved models from baseline B-FT}
\label{table:best_3_ranked_models}
\end{table}

\begin{table}[htbp]
\centering
\resizebox{\columnwidth}{!} {%
\small
\setlength\tabcolsep{2.5pt}
\begin{tabular}{|c|l|c|c|c|c|c|c| }
\hline
Ensemble &  Models  &$Si \rightarrow Ta$ &$Ta \rightarrow Si$ & $Si \rightarrow En$ & $En \rightarrow Si$ & $Ta \rightarrow En$ & $En \rightarrow Ta$\\
\hline
No  & B-FT (Baseline)     & 29.75         & 37.57         & 37.72         & 36.11         & 33.36          & 25.80 \\
\cline{2-8}
Ensemble & Rank-1 & 31.16 (+1.41)  & 39.40 (+1.83)  & 38.89 (+1.17) & 36.91 (+0.8) & 35.96 (+2.6) & 26.83 (+1.03)  \\
\hline
Checkpoint  & B-FT (Baseline)    & 30.07 & 37.86 & 37.77 & 36.21 & 33.40 &	25.82\\
\cline{2-8}
Ensemble & Rank-1   & 31.26 & 39.30 & 39.58 & 36.85 & 34.54 & 26.76\\

 \hline
 & B-FT \& Rank-1  & 31.41 &	39.48 &	38.97 &	37.42 &	35.19 &	26.77  \\
 & B-FT \& Rank-2  & 30.93 &	38.75 &	38.68 &	37.17 &	34.48 &	27.12  \\
 & B-FT \& Rank-3  & 31.16 &	38.80 & 38.26 &	37.54 &	34.65 &	26.57  \\
 & B-FT \& Rank-1,2  & 31.89 (+2.14 ) &	39.31 &	39.93 (+2.21) &	37.69 &	35.55 &	27.82 (+2.02) \\
Multi- & B-FT \& Rank-1,3  & 31.61 	& 39.68 (+2.11) &	39.70 &	37.89 (+1.78) &	35.47 &	27.23  \\
Model & B-FT \& Rank-2,3  &  31.57 &	39.11 &	39.13 &	37.87 &	35.04 &	27.09 \\
Ensemble &  Rank-1,2 &  31.90 &	39.52 &	38.81 &	36.97 &	35.88 (+2.52) &	27.73   \\ 
 &  Rank-1,3 & 31.46 	& 38.60 &	39.83 &	37.29 &	25.95 &	26.61  \\
 &  Rank-1,2,3 &  31.77 	& 39.24 &	39.68 &	37.57 &	35.75 &	27.52 \\
\hline
\end{tabular}
}
\caption{Ensemble results for all the six directions.}
\label{table:ensemble}
\end{table} 

We select the top 3 well-performed models for each direction and rank them accordingly as shown in Table~\ref{table:best_3_ranked_models}. Then we empirically evaluate all the possible combinations of multi-model ensembling and report the results as shown in Table~\ref{table:ensemble}. The checkpoints ensemble slightly outperforms the baseline single model in all directions. As we can see, the multi-model ensemble overall improves the performance in all the translation tasks except on $Ta \rightarrow En$, where a single model stands out. We observe that baseline B-FT also stands as a strong model on multi-model ensemble along with our best-ranked proposed models for the rest of the 5 directions. 
Finally, we obtain a maximum +2.21 BLEU point improvement on Si-En and a minimum of +1.78 BLEU score improvement on En-Si directions relative to our strong baseline single-stage fine-tuning(B-FT).

\section{Discussion}
\label{sec:discussion}


We explore the effectiveness of mixed precision training, the application of CPT and ITTL techniques, and the role of in- and out-domain data in enhancing translation quality across multiple language pairs. The insights derived from these evaluations offer valuable directions for optimizing translation models, particularly in resource-constrained settings. Findings are:
(1). Mixed precision training, while delivering similar BLEU scores as full precision training, significantly enhances training efficiency by reducing training time and memory requirements. This allows for larger batch sizes and increased model capacity without compromising translation accuracy.
(2). CPT approach, especially for the Si - En language pair, shows notable improvements in translation quality over baseline models. The biCPT approach, utilizing in-domain monolingual data (Case A (ii)) followed by bilingual fine-tuning (B-FT), achieved the highest translation quality. In-domain pre-training was found to be more effective (even with limited in-domain monolingual data availability) than strategies relying on out-of-domain or mixed-domain monolingual data. (3). The M2M-FT setup typically underperformed compared to the B-FT baseline due to insufficient parallel data, essential for robust msLLMs. However, the O2M-FT  and M2O-FT configurations generally surpassed M2M-FT in performance (except in three scenarios), aligning with previous results in English-centric MNMT\citep{tang-etal-2021-multilingual}. Specifically, equalizing data volumes between Si - Ta and Si - En significantly improved results for Si - Ta and Ta - Si in the O2M-FT and M2O-FT setups.
(4). The inclusion of in-domain monolingual data in pre-training, combined with the use of both in-domain and out-of-domain data in fine-tuning, is crucial for improving translation performance in underrepresented languages. Although large volumes of out-domain parallel data are beneficial, a substantial volume of in-domain parallel data is essential for capturing effective language representations, particularly in morphologically rich languages.
(5). The baseline B-FT model remains robust, effectively enhancing the performance of multi-model ensembles in five out of six translation directions. Significant improvements were observed in translations between Si and En compared to the baseline model. The findings highlight the general superiority of multi-model ensembles in boosting translation performance across various tasks.

\subsection{Manual Analysis of the Translated Output}

An in-depth manual inspection of the translation outputs for each language direction reveals areas of model limitations. Examples of such translated sentences are included in Appendix~\ref{output_translations}. A recurrent issue is the models' inability to accurately translate place names and roads, highlighting the challenge of domain-specific named entities with limited data coverage. Future enhancements may incorporate data augmentation strategies to integrate these entities. 

A salient observation from our analysis is the diminished performance of our models when translating into Tamil. This reduced efficacy can be attributed to the intrinsic linguistic features of Tamil, such as its flexible word order~\cite{futrell2015quantifying} and more complex inflectional system, compared to Sinhala. The free word-ordering nature of Tamil presents a unique challenge for translation models. Moreover, as indicated by~\cite{anand2010sequence}, Tamil's extensive inflection further exacerbates these challenges. Our scrutiny of the Tamil translations revealed instances of free word order, compound words, and syntactic structures that closely mirror those of the references.

Additionally, the translated outputs often employed words that were synonymous with or similar to those in the reference sentences. This phenomenon underlines a fundamental principle in computational linguistics: as the linguistic complexity of a target language increases, so does the requisite volume of training data needed for a model to capture and reproduce the language's specific features accurately. Consequently, our findings suggest that even when two languages are provided with comparable amounts of monolingual data for pre-training, the complexity of the target language is a determining factor in the success of the fine-tuning process.

\section{Related Work}
\label{sec:lit}

First, we discuss msLLMs and their utility for NMT. Then we analyze CPT and ITTL with msLLMs for NMT. 

\subsection{Multilingual Sequence-to-sequence Large Language Models (msLLMs)}
\label{sec:lit_pre_trained_models}

The BART~\citep{lewis-etal-2020-bart} model is the first complete self-supervised sequence-to-sequence (encoder-decoder) model trained on a large scale (English) monolingual corpus. BART is implemented as a standard Transformer architecture~\citep{10.5555/3295222.3295349} with a bidirectional auto-encoder and a left-to-right autoregressive decoder. Later BART was extended to incorporate more than one language, resulting in mBART25. mBART25~\citep{liu-etal-2020-multilingual-denoising} is the first self-supervised multilingual sequence-to-sequence Large Language Model (msLLM) trained on a large-scale unlabeled monolingual corpus of 25 different languages. ~\cite{tang-etal-2021-multilingual} extended this model upto 50 languages, which resulted in the mBART50 model. Another well-known sequence-to-sequence model is T5~\citep{JMLR:v21:20-074}, a Text-to-Text Transfer Transformer trained over a large English monolingual corpus similar to BART~\citep{lewis-etal-2020-bart}. mT5~\citep{xue-etal-2021-mt5} is the multilingual version of T5, and includes 101 languages~\citep{xue-etal-2021-mt5}. Subsequently introduced mT6~\citep{chi-etal-2021-mt6} has been additionally trained with parallel data on mT5. However, mBART is memory efficient and has shown relatively better results than mT5 for translation~\citep{lee-etal-2022-pre,nayak2023leveraging}.

\subsection{Fine-tuning msLLMs for NMT Tasks}
\label{related_work_ft_msllms}

\begin{figure*}[htp]
    \centering
    \includegraphics[scale=0.5]{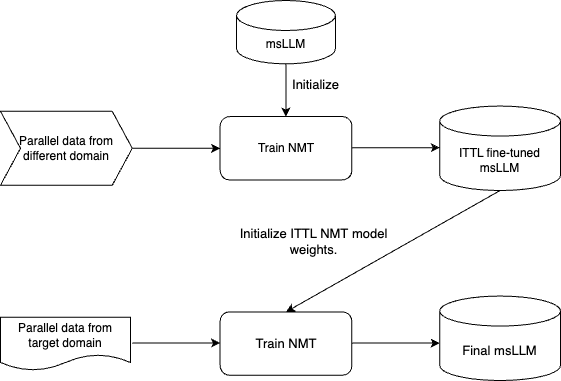}
    \caption{Overview of ITTL}
    \label{fig:ittl}
\end{figure*}

Since msLLMs are trained using a denoising objective, they cannot be directly used as an NMT model. However, when parallel data between a source and target language is available, an msLLM can be used to fully initialize the NMT (encoder-decoder) training, thus avoiding the need to train the NMT model from scratch. This process is referred to as \textit{fine-tuning}. In other words, the parameters of an msLLM can be fine-tuned to incorporate the model in an NMT task in the context of a language pair. Studies show this leads to significant performance gains over NMT models based on the simple Transformer architecture due to the knowledge transfer from msLLMs~\citep{liu-etal-2020-multilingual-denoising,tang-etal-2021-multilingual,thillainathan2021fine}.

In particular, the mBART model has shown promising results on supervised and unsupervised machine translation at both the sentence and document levels~\citep{liu-etal-2020-multilingual-denoising,tang-etal-2021-multilingual}. Significant gains have been observed in LRL pairs such as $English\leftrightarrow Vietnamese/Turkish$ as well~\citep{liu-etal-2020-multilingual-denoising,tang-etal-2021-multilingual}. However, all the experiments conducted by the above research have been English-centric, i.e. English is either on the source or target side of the given language pair. ~\cite{tang-etal-2021-multilingual} experimented with fine-tuning multiple language pairs simultaneously, called Multilingual Fine-Tuning (M-FT). In their experiments also, the tested cases are only English-centric. They experimented with M2O ($Many \rightarrow English$), O2M ($English \rightarrow Many$), and M2M ($Many \leftrightarrow Many$) with English as a pivot language. Some studies considered non-English centric fine-tuning where only one language from the given pair is included in the mBART model~\citep{madaan-etal-2020-transfer,cahyawijaya-etal-2021-indonlg}. ~\cite{thillainathan2021fine} demonstrated that fine-tuning msLLMs for LRLs NMT, particularly in non-English pairs like Tamil to Sinhala, significantly outperforms standard Transformer-based models. Their study also highlighted the importance of the volume of monolingual data in training msLLMs for LRLs, suggesting a need for enhanced pre-training and fine-tuning approaches in LRL-NMT.



\cite{nayak2023leveraging, adelani-etal-2022-thousand} show that ITTL of msLLM models is extremely beneficial for domain-specific NMT, especially when target domain data is limited/unavailable in msLLMs. They analysed the possibility of using auxiliary parallel data to fine-tune the msLLMs before fine-tuning with the limited target domain data (as showed in Figure~\ref{fig:ittl}). Their observation improved the msLLM performance for domain-specific NMT of LRLs and strongly holds for languages missing or under-represented in the msLLM.

Previous studies mainly explored two variants: (1) fine-tune with data from an auxiliary domain (also referred to as intermediate task) and further fine-tune with the target domain \citep{adelani-etal-2022-thousand,nayak2023leveraging} and (2) fine-tune by combining data from different domains and further fine-tune with the target domain~\citep{adelani-etal-2022-thousand}. However, they did not further look into consecutive three-stage fine-tuning strategies and fine-tuning from M-FT models. Building upon the findings of~\cite{nayak2023leveraging, adelani-etal-2022-thousand}, which underscore the significance of ITTL in enhancing the performance of msLLMs, this study delves deeper into the exploration of various strategies for implementing ITTL. 

\subsection{Continual Pre-training of msLLMs (Extending the msLLMs)}
\label{related_work_cpt}

CPT—which entails further pre-training of an LLM with monolingual data—has been found to improve the performance of encoder-based models such as BERT~\citep{devlin-etal-2019-bert} and RoBERTa~\citep{Liu2019RoBERTaAR} in domain-specific text classification tasks~\citep{beltagy-etal-2019-scibert,gururangan-etal-2020-dont,10.1093/bioinformatics/btz682,zhang-etal-2020-multi-stage}. Further studies conducted CPT on parallel corpora~\citep{reid-artetxe-2022-paradise,chi-etal-2021-mt6}, demonstrate that integrating parallel data into msLLMs significantly enhances model performance across various language tasks, outperforming traditional methods that rely mainly on monolingual corpora. Although this continual pre-training strategy has the potential for domain adaptation of msLLMs in NMT, existing research has not adapted it to a particular domain.

CPT has been commonly employed to extend an msLLM to new languages~\citep{{tang-etal-2021-multilingual,liu-etal-2021-continual,chen-etal-2020-facebook,susanto-etal-2021-rakutens}}. 
PARADISE~\citep{reid-artetxe-2022-paradise} extended the conventional pre-training using denoising objective by (i) replacing words in the noised sequence according to a multilingual dictionary, and (ii) predicting the reference translation according to a parallel corpus instead of recovering the original sequence. However, they have not focused on adapting to a particular domain via pre-training on msLLMs.

\section{Conclusion}

This study highlights the effectiveness of refining CPT and ITTL on msLLMs with additional monolingual and parallel data, presenting it as a promising method for LRL-NMT. Previous research primarily English-focused settings, our work broadens these techniques to incorporate non-English-centric translation as well. We developed and assessed CPT and ITTL with various approaches using both in-domain and out-domain parallel data.

Our experimental results indicate that translation accuracy is significantly influenced by both the quantity of monolingual data and its domain relevance to the pre-training phase. Furthermore, we have determined that incorporating even a modest amount of in-domain monolingual data can yield substantial improvements in msLLM performance. Notably, the use of out-domain data in the ITTL fine-tuning process appears to be more beneficial for translations from Sinhala and Tamil into English than the reverse. 

However, our proposed approaches are not specific to one particular msLLMs, they can be applied to generative LLMs as well, leaving that for feature work. Besides, the adaptation of M-FT models has not consistently surpassed the performance of B-FT models. Recognizing this, our future work will aim to refine the M-FT processes. We anticipate that such enhancements will facilitate the development of a singular, comprehensive MNMT model capable of managing all translation directions effectively.  

\begin{acks}
This research was supported by the Accelerating Higher Education Expansion and Development (AHEAD) Operation of the Ministry of Education, Sri Lanka, funded by the World Bank.
\end{acks}

\bibliographystyle{ACM-Reference-Format}
\bibliography{mybibfile}

\appendix
\section{Addressing the Zero Width Joiner (ZWJ) issue}
\label{postprocessing_logic}

ZWJ post-processing fix is mainly required for $En\rightarrow Si$ and $Ta \rightarrow Si$ directions. i.e. when Sinhala is on the target side.  We added the following two lines of code

\begin{lstlisting}
  sentence = sentence.replace("\u0DCA \u0dbb", "\u0DCA\u200D\u0dbb") 
  sentence = sentence.replace("\u0DCA \u0dba", "\u0DCA\u200D\u0dba") 
\end{lstlisting}

 just before the return statement of the following method in Fairseq:   
\begin{lstlisting}
    def post_process(sentence: str, symbol: str):
\end{lstlisting}

\section{Output Translations}
\label{output_translations}
\begin{figure*}[htp]
    \centering
    \includegraphics[scale=0.4]{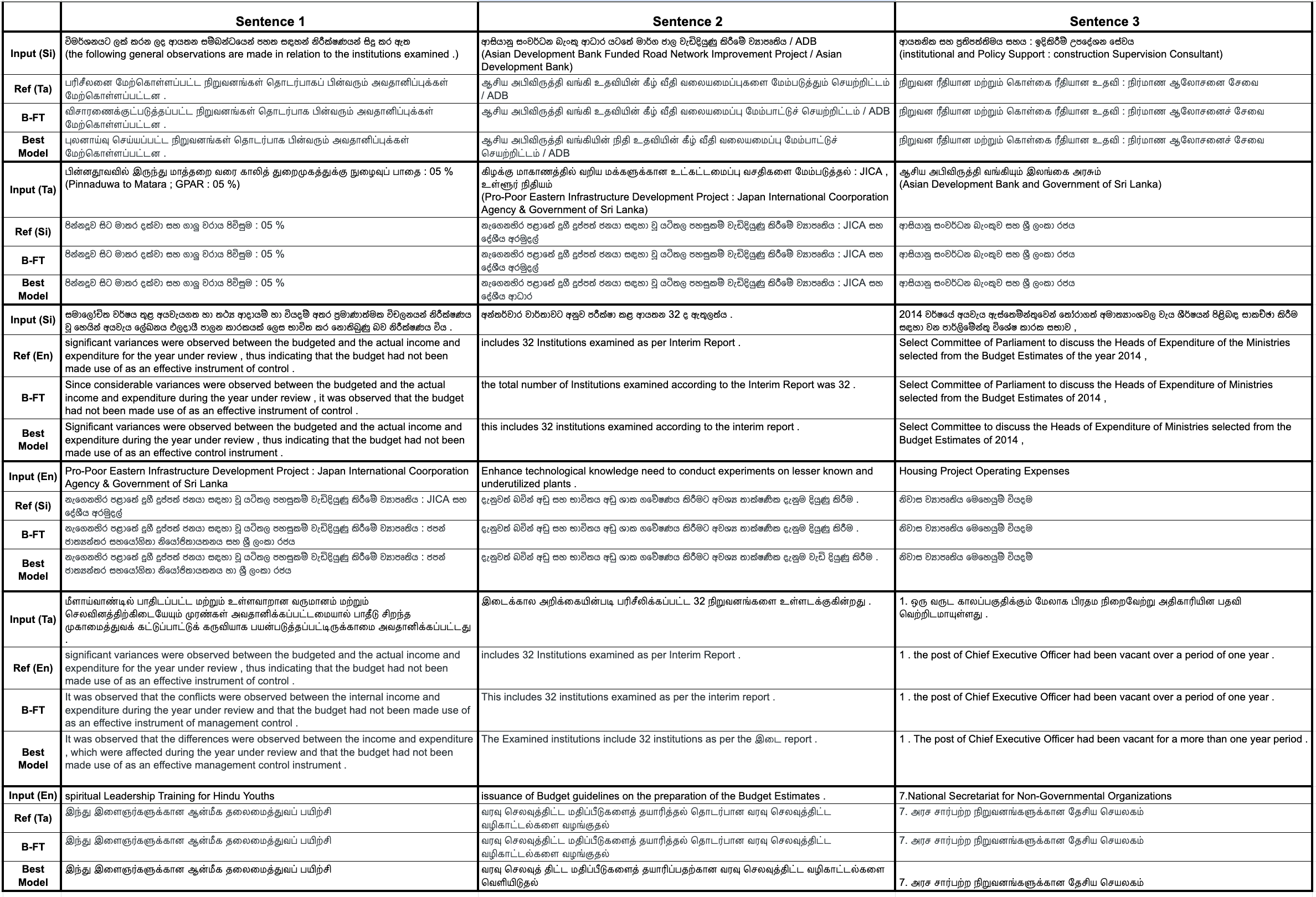}
    \caption{Output translations}
    \label{fig:outputs_translations_examples}
\end{figure*}

Figure~\ref{fig:outputs_translations_examples} highlights some output translations of the best-fine-tuned model against reference texts for all six translation directions.


\end{document}